\documentclass[pmlr]{jmlr}

\usepackage{longtable}

\usepackage{booktabs}
\usepackage[load-configurations=version-1]{siunitx} 

\usepackage{caption}
\usepackage{subcaption}
\usepackage{multirow}
\usepackage{multicol}

\theorembodyfont{\upshape}
\theoremheaderfont{\scshape}
\theorempostheader{:}
\theoremsep{\newline}

\jmlryear{2019}
\jmlrworkshop{Machine Learning for Healthcare}

\title[Alzheimer's Disease Classification Using MRI Scans]{A Comprehensive Study of Alzheimer's Disease Classification Using Convolutional Neural Networks}

\author{
\Name{Ziqiang Guan} \Email{zguan@cs.umass.edu}\\
\Name{Ritesh Kumar} \Email{riteshkumar@cs.umass.edu} \\
\Name{Yi Ren Fung} \Email{yfung@cs.umass.edu} \\
\Name{Yeahuay Wu} \Email{yeahuaywu@cs.umass.edu} \\
\Name{Madalina Fiterau} \Email{mfiterau@cs.umass.edu} \\
\addr College of Information and Computer Science \\
University of Massachusetts Amherst\\
Amherst, MA, U.S.A.
}

\begin{document}

\maketitle


\begin{abstract}
A plethora of deep learning models have been developed for the task of Alzheimer's disease classification from brain MRI scans. Many of these models report high performance, achieving three-class classification accuracy of up to 95\%. However, it is common for these studies to draw performance comparisons between models that are trained on different subsets of a dataset or use varying imaging preprocessing techniques, making it difficult to objectively assess model performance. Furthermore, many of these works do not provide details such as hyperparameters, the specific MRI scans used, or their source code, making it difficult to replicate their experiments. To address these concerns, we present a comprehensive study of some of the deep learning methods and architectures on the full set of images available from ADNI. We find that \textbf{(1)} classification using 3D models gives an improvement of 1\% in our setup, at the cost of significantly longer training time and more computation power, \textbf{(2)} with our dataset, pre-training yields minimal ($<0.5\%$) improvement in model performance, \textbf{(3)} most popular convolutional neural network models yield similar performance when compared to each other. Lastly, we briefly compare the effects of two image preprocessing programs: FreeSurfer and Clinica, and find that the spatially normalized and segmented outputs from Clinica increased the accuracy of model prediction from 63\% to 89\% when compared to FreeSurfer images.
\end{abstract}


\section{Introduction}

Alzheimer's disease (AD) is a degenerative brain disease and the most common cause of dementia. It is a burdensome and costly disease that affects 5.5 million people in the United States. The total value of care provided to AD patients is estimated to be around \$250 billion a year \citep{alzheimer2017facts}. 

The symptoms of AD include memory loss, challenges in planning, difficulty completing simple tasks, and decline in other cognitive skills that often lead to a dependency on external care, which takes a toll on both the family of the individual with AD, as well as society as a whole.

The disease is characterized by the degeneration of specific nerve cells, presence of neuritic plaques outside the neurons, and the accumulation of neurofibillary tangles inside the neurons \citep{mckhann1984clinical}, where the latter two proteins lead to cell death by interfering with various functions of the cell \citep{alzheimer2017facts}. As a result, brains of people with advanced AD show inflammation, dramatic shrinkage from cell loss, and widespread debris from dead or dying neurons \citep{alzheimer2017facts}.

While no cure exists for the disease yet, there is consensus on the need and benefit for early diagnosis of AD. For example, an overwhelming percentage (80\%) of older adults wish to know as early as possible if they have AD \citep{dale2008correlates}, and early diagnosis may help caregivers of AD patients feel more competent in caring for the patients, as well as increasing the possibility that the caregiver can involve the patient in making medical, financial, legal, and care decisions \citep{de2013impact}. Furthermore, \citet{weimer2009early} used Monte Carlo cost-benefit analysis to suggest potential social and financial benefits of early detection. 

Diagnosis of AD is typically done with a comprehensive evaluation of the patient, which may include tests of memory, problem solving, and attention, as well as brain scans using computed tomography (CT), magnetic resonance imaging (MRI), or positron emission tomography (PET) scans. 

Recently, machine learning techniques, specifically deep learning, show great potential in aiding the diagnosis of AD using MRI scans. A variety of neural network architectures have been proposed, some operate on 2D images \citep{hon2017towards} \citep{billones2016demnet} \citep{sarraf2017deepad}, while others operate on 3D MRI scans \citep{wang2018automatic} \citep{hosseini2016alzheimer} \citep{payan2015predicting}. Because of the relatively small number of available MRI scans, many proposed architectures \citep{payan2015predicting} \citep{hosseini2016alzheimer} also incorporate an auto-encoder architecture with tied weights to pre-train the model on a reconstruction task before the classification task to prevent over-fitting on the data.

However, while there is a sizable number of studies on this topic, for people who wish to utilize a discriminitive MRI feature extractor for other downstream AD-related tasks, it is still difficult to glean insights that would be helpful for this task. For example, it is not clear how much performance is gained by using a 3D convolutional neural network (CNN) to process the MRI, which has a high computation cost, compared to using a simple 2D model and processing only one slice, or multiple slices of the MRI. It is also not clear how much pre-training the model helps with raising the performance of the model, as \citet{gupta2013natural}, \citet{payan2015predicting}, \citet{hosseini2016alzheimer} have done. Finally, it is also not clear how much the various model architectures contribute to the high accuracy that various authors have claimed in their paper.

In this study, we explore the aforementioned variables one must consider when selecting a discriminative CNN setup for AD classification, \textbf{1)} the difference in performance between using 2D and 3D models; \textbf{2)} pre-training the model on reconstruction task before training it on the classification task; and \textbf{3)} the performance gain from using model X versus model Y.

\section{Related Work}

The idea of using computers to assist in the diagnosis of AD has been around before the current wave of enthusiasm for neural networks. \citet{plant2010automated} used a feature selection algorithm to achieve 92\% accuracy in a binary classification of AD (Alzheimer's disease) versus HC (Healthy Control). \citet{jha2017diagnosis} used a dual-tree complex wavelet transform to extract features (DTCWT) from the MRI scans, performed principle component analysis (PCA) to reduce the dimension of the extracted features, and a feed-forward neural network to classify between AD and HC individuals. 

Ever since \citet{krizhevsky2012imagenet} won the ImageNet challenge with a seven-layer convolutional neural network, surpassing the previous winner by a large margin of 10\%, the computer vision community has mainly shifted from using hand-crafted features with support vector machines (SVM) or feed-forward neural networks, to large, end-to-end trainable neural networks with many layers.

\citet{gupta2013natural} trained a sparse auto-encoder on both natural images and MRI images to learn a set of bases or filters. The filters were then convolved with the target MRI to produce a set of features, which were then down-sampled using max-pooling before fed into a feed-forward network for classification. Similarly, \citet{payan2015predicting} trained a sparse auto-encoder on randomly selected patches from the MRI scans to learn the features. They also used max-pooling to reduce the size of the features before introducing them into a feed-forward network.


More recently, \citet{hon2017towards} used the image entropy of each slice to decide which of the slices of the MRI scans to use. They then fed the slice through a pre-trained 2D VGG-16 network \citep{simonyan2014very}, as well as Inception V4 network \citep{szegedy2015going}, to obtain the final result.

\citet{wang2018automatic} used DenseNet \citep{huang2017densely} and ensemble methods to classify the entire 3D MRI scan, leading to a new state-of-the-art three-class (Alzheimer versus Mild Cognitive Impairment versus Cognitively Normal) classification accuracy of 97.19\%. Their study showed that around 2.5\% increase in accuracy was attributed to applying ensemble method, which falls in-line with other studies that employed the method.


\section{Methods}

\begin{figure}[htbp]
  \centering 
  \includegraphics[width=4.5in]{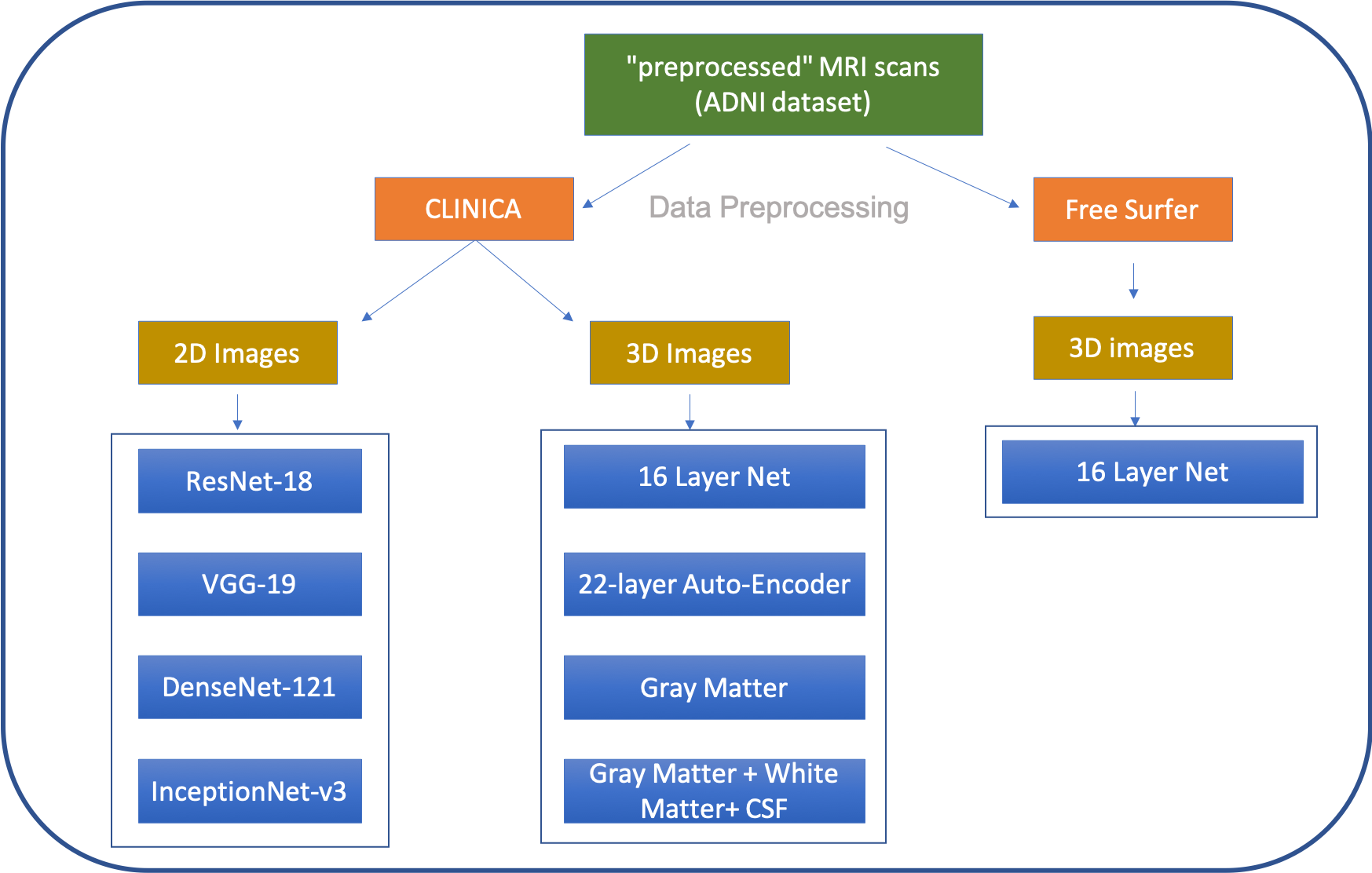} 
  \caption{An overview of the pipeline of our experiments. The ``preprocessed" MRI scans from ADNI are processed by Clinica, then used for both 2D and 3D classification. The FreeSurefer scans from ADNI are used directly for 3D classification.}
  \label{fig:method-overview} 
\end{figure}

In this section, we detail the setup we used for this study, including the acquisition and preprocessing of the data, and the model architectures we use in our comparison. Figure \ref{fig:method-overview} provides an overview of the various components our study will cover.

\subsection{Data Acquisition}

For this study, we obtained MRI scans from the  Alzheimer's Disease Neuroimaging Initiative (ADNI) \footnote{ADNI Website: \href{http://adni.loni.usc.edu/}{http://adni.loni.usc.edu/}}. Specifically, we are using MRI scans with the description ``MPR; GradWarp; B1 Correction; N3", which we will henceforth refer to as ``preprocessed" scans, as well as the FreeSurfer post-processed MRI scans with the description ``FreeSurfer Cross-Sectional Processing brainmask" in the ADNI database. A total of 3415 ``preprocessed" scans and 3177 FreeSurfer post-processed scans were obtained from the website. 



    

\begin{center}
    \begin{figure*}[t]
    \begin{minipage}{\linewidth}
    \centering
    
    \includegraphics[width=0.2\linewidth]{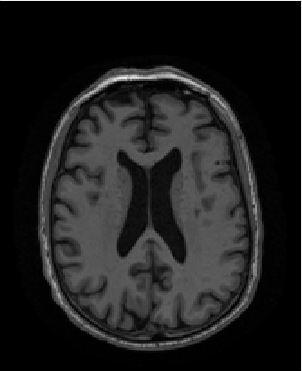}
     \includegraphics[width=0.2\linewidth]{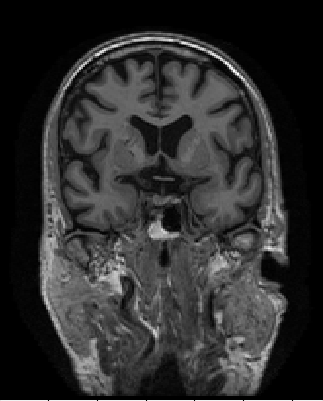}
     \includegraphics[width=0.2\linewidth]{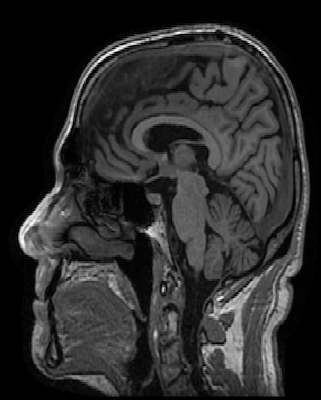}
    \\ \textbf{(a)} The axial, coronal, and sagittal view of the ``preprocessed" scans.
    \vspace{1mm}
    \end{minipage}
    
    \begin{minipage}{\linewidth}
    \centering
    \includegraphics[width=0.2\linewidth]{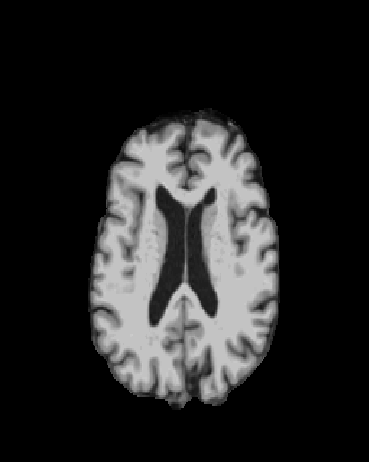}
     \includegraphics[width=0.2\linewidth]{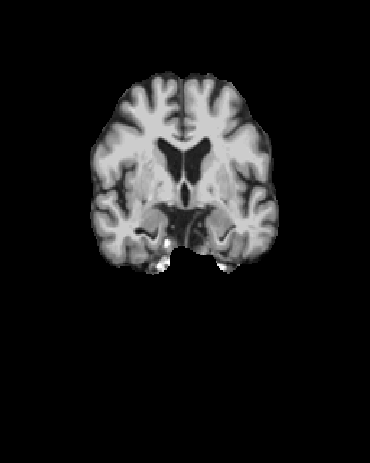}
     \includegraphics[width=0.2\linewidth]{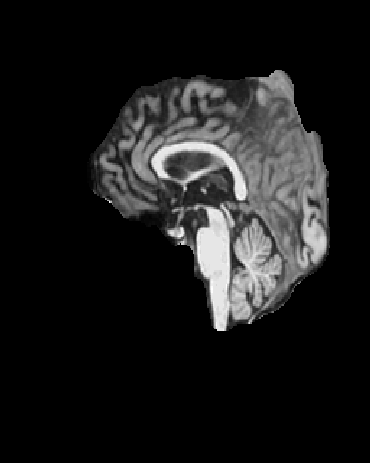}
    \\ \textbf{(b)} The axial, coronal, and sagittal view of the FreeSurfer post-processed scans.
    \vspace{1mm}
    \end{minipage}
    
    \begin{minipage}{\linewidth}
    \centering
    \includegraphics[width=0.2\linewidth]{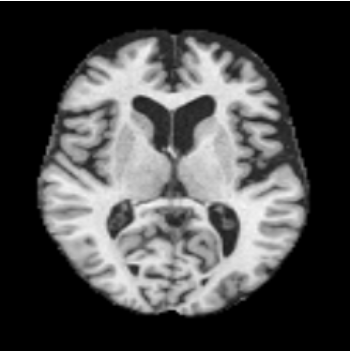}
     \includegraphics[width=0.2\linewidth]{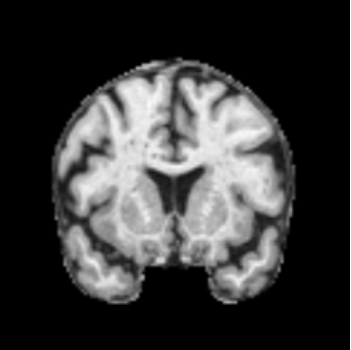}
     \includegraphics[width=0.2\linewidth]{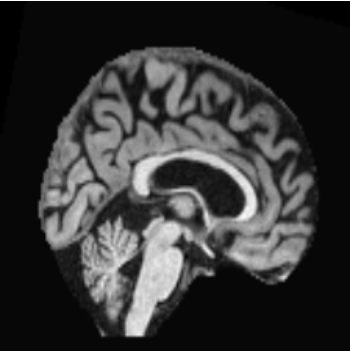}
    \\ \textbf{(c)} The 52nd axial, 92nd coronal, and 58th sagittal view of the brain extraction outputs from Clinica, which we used for 2D classification.
    \vspace{1mm}
    \end{minipage}
    
    \captionof{figure}{Slices from ``preprocessed" scans, FreeSurfer post-processed scans, and the specific slices we use for 2D classification.}
    \label{fig:mri-slices}
    \end{figure*}
\end{center}

\subsection{Data Preprocessing}

Due to the prominence of extraneous parts of the head that are captured in MRI scans, which can potentially intorduce noise in the network, most studies use FSL\footnote{FSL: \href{https://fsl.fmrib.ox.ac.uk/fsl/fslwiki/FSL}{https://fsl.fmrib.ox.ac.uk/fsl/fslwiki/FSL}} and/or Statistical Parameteric Mapping (SPM)\footnote{SPM: \href{https://www.fil.ion.ucl.ac.uk/spm/}{https://www.fil.ion.ucl.ac.uk/spm/}} for the preprocessing of the MRI scans. FSL provides brain extraction and tissue segmentation functionality, and SPM realigns, spatially normalizes, and smoothes the scans.

For our experiments, we used the Clinica software platform\footnote{Clinica: \href{http://www.clinica.run}{http://www.clinica.run}} developed by the ARAMIS Lab\footnote{Aramis Lab: \href{www.aramislab.fr}{www.aramislab.fr}}, which supports FSL, SPM, FreeSurfer, as well as a few other technologies. We are using the t1-volume pipeline, which is a wrapper of the ``segmentation", ``run dartel", and ``normalize to mni space" routines implemented in SPM.

The inputs to the Clinica pipeline were the ``preprocessed" scans from ADNI, and the outputs of the Clinica pipeline include spatially normalized gray matter, white matter, and cerebrospinal fluid segmentation maps, as well as a brain extraction output similar to the FreeSurfer post-processed scans from ADNI's dataset, with a difference being that the Clinica outputs are spatially normalized. 

Furthermore, we extract 2D slices from fixed indices for each view of the spatially normalized brain extraction output. The indices were chosen based on visual prominence of the hippocampus of the brain. Specifically, they were the 52nd slice of axial view, the 58th slice of sagittal view, and the 92nd slice of coronal view. We also experimented with using neighboring slices for classification. Figure \ref{fig:mri-slices} shows the 2D slices of the brain extraction output by Clinica. 

\subsubsection{Demographics}

After processing the ``preprocessed" scans with Clinica, we obtained 2023 sets of scans. Table \ref{table:freesurfer-demo} shows the demographics of the FreeSurfer post-processed scans, and Table \ref{table:clinica-demo} shows the demographics of the outputs from the Clinica pipeline. Note that the output from Clinica had fewer scans because some were lost in the process due to unknown reasons.

\begin{table}[h!]
\centering
\begin{center}
\begin{tabular}{ |c|c|c|c|c|c|c| } 
\hline
\multirow{2}{4em}{Class} & \multirow{2}{4em}{No. of subjects} & 
\multirow{2}{4em}{No. of scans} & \multicolumn{2}{|c|}{Sex} & \multicolumn{2}{|c|}{Age}\\
\cline{4-7} & & & Male & Female & Mean & Std. Dev\\
\hline
AD & 326 & 928 & 185 & 141 & 74.8 & 7.3\\  \hline
MCI & 219 & 1408 & 141 & 78 & 75.1 & 5.3\\  \hline
CN & 217 & 982 & 118 & 99 & 75.6 & 7.3\\
\hline
\end{tabular}
\end{center}
\caption{Demographics for the FreeSurfer post-processed scans from ADNI's website.}
\label{table:freesurfer-demo}
\end{table}


\begin{table}[h!]
\centering
\begin{center}
\begin{tabular}{ |c|c|c|c|c|c|c| } 
\hline
\multirow{2}{4em}{Class} & \multirow{2}{4em}{ No. of subjects} & 
\multirow{2}{4em}{No. of scans} & \multicolumn{2}{|c|}{Sex} & \multicolumn{2}{|c|}{Age}\\
\cline{4-7}& & & Male & Female & Mean & Std. Dev\\
\hline
AD & 199 & 586 & 106 & 93 & 75.4 & 7.3\\ \hline
MCI & 237 & 827 & 144 & 93 & 76.1 & 5.4\\ \hline
CN & 169 & 610 & 81 & 88 & 75.4 & 7.1\\
\hline
\end{tabular}
\end{center}
\caption{Demographics for the output scans of the Clinica pipeline. Note that some of the scans were lost in the process due to unknown reasons.}
\label{table:clinica-demo}
\end{table}

\subsubsection{Training / Validation / Testing Split}
\label{sec:set-split}

To evaluate the performance of our models, we split the data into a training set, validation set, and testing set, with a ratio of 6:2:2. We save the model with the lowest validation loss, and use that to obtain the final accuracy on the test set.

\begin{table}[h!]
\centering
\begin{center}
\begin{tabular}{ |c|c|c|c|c|c| } 
\hline
& &  AD    &    MCI   &   CN   &   Total   \\
\cline{2-6}
\multirow{4}{5em}{Clinica}  & Train & 330 & 330 & 330 & 990\\
\cline{2-6} & Validation & 110 & 110 & 110 & 330\\
\cline{2-6}& Test & 110 & 110 & 110 & 330\\
\hline

\multirow{3}{5em}{FreeSurfer}& Train & 351 & 351 & 351 & 1053\\
\cline{2-6}& Validation & 117 & 117 & 117 & 351\\
\cline{2-6}& Test & 117 & 117 & 117 & 351\\
\hline
\end{tabular}
\end{center}
\caption{Data distribution between the three classes (AD versus MCI versus CN) and training, validation, and testing folds for both FreeSurfer post-processed scans and Clinica outputs.}
\label{table:data-distribution}
\end{table}

For each run, we split the data into three classes: (AD, MCI, CN). We then perform shuffling on each of the splits. The first $N$ scans of each class are used to create a balanced dataset, where $N$ is the size of the smallest class. We then split each of the classes into training, validation, and testing set. See Section \ref{sec:split-method} in Appendix A for details on the procedure, and Table \ref{table:data-distribution} for the final distribution for both the Clinica and FreeSurfer images.

Many previous works such as \citet{wang2018automatic}, \citet{hosseini2016alzheimer}, and \citet{sarraf2017deepad}, performed ten-fold cross-validation as a way of measuring the performance of their model. We decided to deviate from this setup because we are not performing hyperparameter search, and therefore require an extra set of data for final evaluation so we can use the validation set to guide early-stopping. To compensate for the reduced number of possible combinations of data that we can train, validate, and test in one run, we utilize the aforementioned shuffling scheme, and run each experiment four to six times to obtain a good randomized sample over the training, validation, and testing set. All of the 2D experiments were run six times, but we were only able to run the 3D experiments four times each due to the significantly longer time and greater computation power required.

    

\subsection{2D Convolutional Neural Network Models}

For classifying 2D slices of MRIs, we use an 18-layer residual network \citep{he2016deep}, a 16-layer VGG network \citep{simonyan2014very}, a 121-layer DenseNet \citep{huang2017densely}, and the Inception V3 architecutre \citep{szegedy2016rethinking}. We believe that our selection of networks is representative of the most popular network architectures currently being used by the computer vision community, and the designs of many of the more complex networks we have come across, such as \citet{wang2018automatic}'s 3D DenseNet, borrow architectural ideas from these networks.

The models we selected are pre-built models from the PyTorch model zoo, and unless specified otherwise, the weights of the models are initialized from weights pre-trained on the ImageNet dataset. The last layer of each of the models was replaced with a linear layer that reduced the dimension of the extracted features of the convolutional layers down to three, each corresponding to one of three classes (AD/MCI/CN).

\subsection{3D Convolutional Neural Network Models}
\label{sec:method-3d-cnn}

Because there are no pre-built 3D models available, we built a 16-layer (Table \ref{tab:16-layer-network}) and 22-layer (Table \ref{tab:22-layer-network}) CNN model following the architectural designs of the residual network \citep{he2016deep}. Specifically, we used the ``bottleneck" configuration to reduce the number of filters in the inner layer of each residual block \citep{he2016deep}, and the ``full pre-activation" layout \citep{he2016identity} for the residual blocks. The layout of each residual block can be found in Table \ref{table:res-block}.

\begin{table}[]
    \centering
    \begin{tabular}{c|c|c|c|c}
        & In Channel & Out Channel & Kernel Size & Stride \\ \hline
        convolution layer & 1 & 32 & 3 & 1 \\ \hline
        residual block & 32 & 32 & --- & --- \\ \hline
        convolution layer & 32 & 64 & 3 & 2 \\ \hline
        residual block & 64 & 64 & --- & --- \\ \hline
        convolution layer & 64 & 128 & 32 & \\ \hline
        residual block & 128 & 128 & --- & --- \\ \hline
        convolution layer & 128 & 256 & 3 & 2 \\ \hline
        convolution layer & 256 & 512 & 3 & 2 \\ \hline
        convolution layer & 512 & 512 & 3 & 2 \\ \hline
        convolution layer & 512 & 512 & 3 & 1 \\ \hline
        linear layer & 32768 & 3 & --- & ---
    \end{tabular}
    \caption{Architecture and hyperparameters for the 16-layer 3D residual network.}
    \label{tab:16-layer-network}
\end{table}

\begin{table}[]
    \centering
    \begin{tabular}{c|c|c|c|c}
        & In Channel & Out Channel & Kernel Size & Stride \\ \hline
        convolution layer & 1 & 32 & 3 & 1 \\ \hline
        residual block	& 32 & 32 & --- & --- \\ \hline		
        convolution layer & 32 & 64 & 3 & 2 \\ \hline
        residual block	& 64 & 64 & --- & --- \\ \hline 	
        convolution layer & 64 & 128 & 3 & 2 \\ \hline
        residual block & 128 & 128 & --- & --- \\ \hline		
        convolution layer & 128 & 256 & 3 & 2 \\ \hline
        residual block & 256 & 256 & --- & --- \\ \hline		
        convolution layer & 256 & 512 & 3 & 2 \\ \hline
        residual block & 512 & 512 & --- & --- \\ \hline		
        convolution layer & 512 & 512 & 3 & 2 \\ \hline
        convolution layer & 512 & 512 & 3 & 1 \\ \hline
        linear layer & 32768 & 3 & --- & --- \\
    \end{tabular}
    \caption{Architecture and hyperparameters for the 22-layer 3D residual network.}
    \label{tab:22-layer-network}
\end{table}


\begin{table}
    \centering
        \begin{center}
            \begin{tabular}{ c|c|c|c|c }
                & In Channel & Out Channel & Kernel Size & Stride \\
                \hline
                Batch Norm & --- & --- & --- & --- \\
                \hline
                ReLU & --- & --- & --- & --- \\
                \hline
                Convolution Layer & n & n/2 & 1 & 1 \\
                \hline
                Batch Norm & --- & --- & --- & --- \\
                \hline
                ReLU & --- & --- & --- & --- \\
                \hline
                Convolution Layer & n/2 & n/2 & 3 & 1 \\
                \hline
                Batch Norm & --- & --- & --- & --- \\
                \hline
                ReLU & --- & --- & --- & --- \\
                \hline
                Convolution Layer & n/2 & n & 1 & 1 \\
            \end{tabular}
        \end{center}
    \caption{The layers for the residual block used in our 3D residual network. n is the number of input channels.}
    \label{table:res-block}
\end{table}

\subsubsection{Convolutional Autoencoder}

Similar to \citet{hosseini2016alzheimer} and \citet{payan2015predicting}, we added a decoder to our 22-layer CNN by reversing the weights of the CNN to form an autoencoder. Specifically, a simplified formula for the operation is described in Equation \ref{eq:weight-tying}, where $X$ is the input matrix, $W$ is the weight matrix, $f$ is an activation function, $Y$ is a matrix representing the feature extracted from $X$, $\hat{X}$ is a reconstruction of $X$ by the network, and $\cdot$ denotes dot product.

\begin{equation}
    \begin{aligned}
        Y & = f(W^T \cdot X) \\
        \hat{X} & = f(W \cdot Y)
    \end{aligned}
    \label{eq:weight-tying}
\end{equation}

\subsection{Optimization Process}

\subsubsection{Reconstruction Task}

To train the convolutional autoencoder on the reconstruction task, we minimize the mean squared error (MSE), its formulation is shown in Equation \ref{eq:mse}, where $X$ is the input matrix (the MRI scan), $\hat{X}$ is obtained from the formulation shown in Equation \ref{eq:weight-tying}, $n$ is the number of MRI scans, and $m$ is the number of features (or voxels) in the scan.

\begin{equation}
    MSE = \frac{1}{n} \sum^{n}_{i=1} \sum^{m}_{j=1} (X_{ij} - \hat{X}_{ij})^2
    \label{eq:mse}
\end{equation}

Following \citet{hosseini2016alzheimer} and \citet{payan2015predicting}, we apply a sparsity constraint $\beta$ to the hidden representation $Y$, giving us the end-to-end formulation of the objective for the reconstruction task, shown in Equation \ref{eq:ete-reconstruction}. We exclude the formulation in Equation \ref{eq:ete-reconstruction} for the weight decay (L2 Regularization) for brevity.

\begin{equation}
    \begin{aligned}
         Loss = \frac{1}{n} \sum^{n}_{i=1} \sum^{m}_{j=1} (X_{ij} - \hat{X}_{ij})^2 + \beta * Y_{ij}
        \label{eq:ete-reconstruction}
    \end{aligned}
\end{equation}

\subsubsection{Classification Task}

For the classification task, we minimize the cross-entropy loss between the predicted probability distribution of the correct label and the ground truth distribution. The objective has the following formulation, shown in Equation \ref{eq:cross-entropy-loss}, where $\hat{p}$ is a vector of probability distribution predicted by the model, $l$ is index of the correct class (e.g. 0 for AD, 1 for MCI, 2 for CN), and $\hat{p}_l$ and $\hat{p}_j$ are the $l$th and $j$th element of $\hat{p}$, respectively.

\begin{equation}
    Loss(\hat{p}, l) = - \log ( \frac{\exp(\hat{p}_l)}{\sum^{\# classes}_{j=1} \exp(\hat{p}_j)} )
    \label{eq:cross-entropy-loss}
\end{equation}

\subsubsection{Training}
\label{sec:training}

We ran all of our experiments for 100 epochs. The weights that led to the lowest validation loss were saved and later used to make predictions on the test set. We chose to save the model with the lowest validation loss instead of the lowest validation accuracy because the validation loss captures the divergence between the predicted distribution and the true distribution, whereas accuracy may or may not capture this.

For hardware, hyperparameters, and other information on training, see Section \ref{sec:addtl-train} in Appendix A.

\section{Results}

\subsection{Evaluation Approach} 

The goal of this study is to answer three queries one may have when deciding among the numerous proposed architectures as feature extractors for other AD-related downstream tasks. Specifically, we are looking to understand the performance gain, if any, from \textbf{1)} using a 2D model on a slice of the scan versus 3D model on the entire MRI scan, \textbf{2)} pre-training the model on the reconstruction task before training it on the classification task, and \textbf{3)} different choices of models and architectures.

We attempt to provide insight into these three questions by keeping all other variables constant, such as preprocessing pipelines, variations in data distribution, hyperparameters for training and testing.

All of the experiments were performed at least four times, and no more than six times, due to resource constraints. To calculate the reported validation accuracy, we take the highest validation accuracy that we encounter in every run, and take the mean as the final validation accuracy. For the test accuracy. we simply take the mean of all of the test accuracy across the runs.

\subsection{2D versus 3D Models} 

For this comparison, we evaluated the performance of 2D ResNet-18 pre-trained on ImageNet, and our own 3D ResNet-22 pre-trained on extra MRI scans left over from class balancing (see Section \ref{sec:set-split}). The 2D ResNet-18 is trained on the coronal slice of the gray matter map from Clinica, and the 3D ResNet-18 is trained on the entire gray matter map. 

From Table \ref{tab:2d-vs-3d}, we can conclude that there is a performance gain of close to 1\% from utilizing 3D models instead of similar 2D models, which may be desirable if accuracy is crucial in the target application. However, there is a large cost of going from 2D to 3D, namely the time required to train the model, and the large amount of GPU memory required to fit the image and model. See Section \ref{sec:addtl-train} in Appendix A for more training information.

\begin{table}[h!]
  \centering 
  \begin{tabular}{c|c|c}
    Model name & Validation Accuracy & Test Accuracy \\ \hline
    2D gray matter (coronal) & $89.5 \pm 1.8\%$ & $87.4 \pm 2.0\%$ \\ \hline
    3D gray matter - w/ pre-train & $90.3 \pm 1.2\%$ & $88.5 \pm 1.7\%$ \\
  \end{tabular}
  \caption{A comparison between 2D ResNet-18 pre-trained on ImageNet and 3D ResNet-22 pre-trained on MRI scans. There is a slight performance advantage with the 3D model. } 
  \label{tab:2d-vs-3d} 
\end{table}

\subsection{Pre-training versus No Pre-Training} 

For this comparison, we evaluate the effects of pre-training on the model's ability to learn and the final accuracy. Table \ref{tab:pre-train-vs-no-pre-train} contains the final accuracy, where there is little evidence to suggest that the model is unable to learn without pre-training, or that pre-training leads to better performance. 


\begin{table}[h!]
  \centering 
  \resizebox{\textwidth}{!}{
  \begin{tabular}{c|c|c}
    Model & Validation Accuracy & Test Accuracy  \\ \hline
    gray matter - no pre-train & $90.1 \pm 1.0\%$ & $88.9 \pm 2.0\%$ \\ \hline 
    gray matter - w/ pre-train & $90.3 \pm 1.2\%$ & $88.5 \pm 1.7\%$ \\ \hline 
    gray matter + CSF + white matter - no pre-train & $91.0 \pm 0.7 \%$ & $89.2 \pm 1.9 \%$ \\ \hline 
    gray matter + CSF + white matter - w/ pre-train & $90.4 \pm 0.8\%$ & $88.9 \pm 2.0\%$ \\
  \end{tabular}
  }
  \caption{Comparison between models with pre-training versus models without pre-training. The results show negligible difference in performance, with a difference in mean accuracy of less than $< 0.5\%$.}.
  \label{tab:pre-train-vs-no-pre-train} 
\end{table}

\subsection{Model Comparisons}

For this comparison, we look at how much impact the model architecture influences the performance of the model, by comparing four popular 2D networks. Table \ref{tab:model-comparison} shows a comparison of the results.

Again, we observe little difference between the various architectures, suggesting that the commonly used architectures should be able to perform well on the task.


\begin{table}[h!]
  \centering 
  \begin{tabular}{c|c|c}
    Model name & Validation Accuracy & Test Accuracy  \\ \hline
    ResNet-18 (coronal) & $89.2 \pm 2.5\%$ & $87.5 \pm 1.1\%$ \\ \hline
    DenseNet121 (coronal) & $90.0 \pm 1.3 \%$ & $88.8 \pm 1.9 \%$ \\ \hline
    InceptionNet\_V3 (coronal) & $89.1 \pm 1.7 \%$ & $88.6 \pm 1.4 \%$ \\ \hline 
    VGG19\_bn (coronal) & $89.1 \pm 1.6 \%$ & $88.1 \pm 2.7 \%$ \\
  \end{tabular}
  \caption{Three-class classification accuracy of the 2D models.} 
  \label{tab:model-comparison} 
\end{table}

\subsection{FreeSurfer versus Clinica}

We performed an additional set of experiments to gain some insight into the difference in performance between using FreeSurfer data from the ADNI website, which were skull-stripped but not spatially normalized, and white matter, gray matter, and cerebrospinal fluid maps from the Clinica pipeline. 

We used the same 3D 16-layer ResNet that we built for both experiments, and found a significant difference in accuracy between the two sets, highlighting the importance and advantage of using the Clinica pipeline. See Table \ref{tab:clinica-vs-freesurfer} for the accuracy comparison.

\begin{table}
  \centering 
  \begin{tabular}{c|c|c}
    Preprocessing Pipeline & Validation Accuracy & Test Accuracy \\ \hline
    FreeSurfer & $77.2 \pm 0\%$ & $63.5 \pm 0\%$ \\ \hline
    Clinica & $91.4 \pm 1.2\%$ & $89.2 \pm 0.4\%$ \\
  \end{tabular}
  \caption{Comparison between FreeSurfer data, which does not have spatial normalization, among other preprocessing procedures, and the Clinica output data. Both experiments were run with our 3D 16-layer ResNet. } 
  \label{tab:clinica-vs-freesurfer} 
\end{table}

\section{Discussion}
By comparing the various setups, we have come to the conclusion that: \textbf{(1)} there is a slight improvement of about 1\% in performance by using a 3D model over a 2D model, \textbf{(2)} the advantage of pre-training the model on MRI scans before training it end-to-end for classification was not noticeable, \textbf{(3)} all of the current popular 2D CNN architectures had similar performance.

For \textbf{(1)}, we leave the reader with the caveat that training a 3D model requires significantly more time and computation power, and the extra 1\% gain in accuracy may not be worth the cost.

For \textbf{(2)}, while we have concluded that pre-training does not aid our model in achieving higher accuracy, we are not rebutting the previous studies over their claim that pre-training helped. We have to consider that we have more data available to us, likely enough to a point where pre-training may no longer be necessary.

For \textbf{(3)}, our results suggest that the features of the model, such as densely connected layers or Inception modules, do not contribute as much to the performance in the task of AD classification as previous studies \citep{wang2018automatic}\citep{hon2017towards} had suggested.

\bibliography{ad_classification}

\appendix
\label{app:A}
\section*{Appendix A.}


\subsection{Data Splitting Methodology}
\label{sec:split-method}

Suppose we have 20 scans, where 7 are labeled as AD, 5 are labeled as MCI, and 8 are labeled as CN. We first shuffle each of the three splits, then take the first 5 scans from each of the class sets, where $5 = \min(7, 5, 8)$, resulting in a final dataset size of 15 from the three classes. We then split each of the three sets into training, validation, and testing sets, resulting in nine total sets. Lastly, we combine the training, validation, and testing sets across all three classes to create a final training set of 9 scans, validation set of 3 scans, and testing set of 3 scans.

\subsection{Additional Training Information}
\label{sec:addtl-train}

\subsubsection{Hardware and Training Time}
For training the 3D models, we used pairs of NVIDIA Tesla M40 GPUs with 24GB of memory. We were able to fit two images on each of the GPUs, bringing our maximum possible batch size to 4. The training time for the experiments ranged from 12-14 hours for the Clinica preprocessed images, to 24-26 hours for the FreeSurfer images. The significantly longer training time was mostly due to the larger file size, which led to slow network IO and longer computation time.

For training the 2D models, we used NVIDIA Titan X GPUs with 12GB of memory. We used a batch size of 16, which was not dictated by memory limit, but rather by intuition, as smaller batch size intuitively leads to better generalization with adaptive learning algorithms such as Adam. The training time for most 2D models completed within 2 hours.

\subsubsection{Hyperaparameters}
For all of the 3D models, we used Adam optimizer for both the reconstruction task and the classification task. For the reconstruction task, we used a learning rate of 0.001 and weight decay (also known as L2 regularization) of 0.0001, as well as 0.0003 for $\beta$, the sparsity constraint parameter. For the classification task, we use a learning rate of 0.001 and weight decay (L2 regularization) of 0.0001. 

For the 2D models, we used Adam optimizer for the classification task, with learning rate set to 0.001 and weight decay set to 0.01 for ResNet18, and learning rate set to 0.0001 and weight decay set to 0.01 for Inception V3, VGG-16, and DenseNet.

The aforementioned hyperparameters are chosen from past experience, as they generally work well for Adam. To keep the experiments consistent, no additional hyperparameter search was performed, with the exception of training Inception V3, VGG-16, and DenseNet, where we lowered the learning rate from 0.001 to 0.0001 because the models were having difficulty training with the higher learning rate.

Table \ref{tab:hyperparameters} lists each of the training hyperparameters we used for the different models and tasks.

\subsection{Tables}

\begin{table}[h]
  \centering 
  \begin{tabular}{|c|c|c|c|}\hline
    Model name & Validation Accuracy & Test Accuracy & No. MRI Data \\ \hline
    ResNet-18 (coronal) & $89.2 \pm 2.5\%$ & $87.5 \pm 1.1\%$ & $2023$ \\ \hline
    ResNet-18 (sagittal) & $88.65 \pm 1.9 \%$ & $88.0 \pm 2.5 \%$ & $2023$ \\ \hline
    ResNet-18 (axial) & $89.7 \pm 3.2\%$ & $87.5 \pm 2.5\%$ & $2023$ \\ \hline
    ResNet-18 (coronal) gray matter only & $89.5 \pm 1.8\%$ & $87.4 \pm 2.0\%$ & $2023$ \\ \hline
    DenseNet121 (coronal) & $90.0 \pm 1.3 \%$ & $88.8 \pm 1.9 \%$ & $2023$ \\ \hline
    InceptionNet\_V3 (coronal) & $89.1 \pm 1.7 \%$ & $88.6 \pm 1.4 \%$ & $2023$ \\ \hline 
    VGG19\_bn (coronal) & $89.1 \pm 1.6 \%$ & $88.1 \pm 2.7 \%$ & $2023$ \\ \hline
    \citet{billones2016demnet} & --- & $91.9\%$ & $900$ \\ \hline
    \citet{hon2017towards} VGG-16 & --- & $92.3 \pm 2.4\%$ & $200$ \\ \hline
    \citet{hon2017towards} Inception V4 & --- & $96.25 \pm 1.2\%$ & $200$ \\ \hline
  \end{tabular}
  \caption{Three-class classification accuracy of our 2D models, as well as a few other studies for comparison.} 
  \label{tab:2d-results} 
\end{table}

\begin{table}[h]
  \centering 
  \resizebox{\textwidth}{!}{
  \begin{tabular}{|c|c|c|c|}\hline
    Model & Val Accuracy & Test Accuracy & \# Subjects \\ \hline
    gray matter - no pre-train & $90.1 \pm 1.0\%$ & $88.9 \pm 2.0\%$ & $605$ \\ \hline 
    gray matter - w/ pre-train & $90.3 \pm 1.2\%$ & $88.5 \pm 1.7\%$ & $605$ \\ \hline 
    gray matter + CSF + white matter - no pre-train & $91.0 \pm 0.7 \%$ & $89.2 \pm 1.9 \%$ & $605$ \\ \hline 
    gray matter + CSF + white matter - w/ pre-train & $90.4 \pm 0.8\%$ & $88.9 \pm 2.0\%$ & $605$ \\ \hline
    
    \citet{wang2018automatic} (w/o ensemble) & --- & 94.7\% & 833 \\ \hline
    \citet{hosseini2016alzheimer} & --- & 94.8\% & 210 \\ \hline
    \citet{jain2019convolutional} & --- & 95.7\% & 150 \\ \hline
    \citet{payan2015predicting} & --- & 89.4\% & 755 \\ \hline
    \citet{khvostikov20183d} (MRI only) & --- & 85.4\% & 200 \\ \hline
  \end{tabular}
  }
  \caption{Three-class classification accuracy for our 3D models, as well as a few other studies for comparison. Our models in this table is the 22-layer residual network, which was mentioned in Section \ref{sec:method-3d-cnn}.}
  \label{tab:3d-results} 
\end{table}

\begin{table}[h]
    \centering
    \begin{tabular}{c|c|c|c|c|c}
        & task & optimizer & learning rate & weight decay & sparsity param \\
        \hline
        ResNet-16 (3D) & R & Adam & 0.001 & 0.0001 & 0.0003 \\
        \hline
        ResNet-22 (3D) & R & Adam & 0.001 & 0.0001 & 0.0003 \\
        \hline
        ResNet-16 (3D) & C & Adam & 0.001 & 0.0001 & --- \\
        \hline
        ResNet-22 (3D) & C & Adam & 0.001 & 0.0001 & --- \\
        \hline
        ResNet-18 (2D) &  C & Adam & 0.001 & 0.01 & --- \\
        \hline
        Inception V3 (2D) & C & Adam & 0.0001 & 0.01 & --- \\
        \hline
        VGG-16 (2D) & C & Adam & 0.0001 & 0.01 & --- \\
        \hline
        DenseNet (2D) & C & Adam & 0.0001 & 0.01 & --- \\
    \end{tabular}
    \caption{The training hyperparameters used for all of the models. In the task column, R denotes reconstruction, and C denotes classification.}
    \label{tab:hyperparameters}
\end{table}

\end{document}